\documentclass[letterpaper, 10 pt, journal, twoside]{ieeetran} 
\IEEEoverridecommandlockouts                              

\usepackage[centertags]{amsmath}

\usepackage{subfigure}
\usepackage{enumitem}

\usepackage[svgnames]{xcolor}
\usepackage[colorlinks=true, linkcolor=black, urlcolor=black, citecolor=black]{hyperref}

\usepackage{pdflscape}

\usepackage{pdfpages} 
\usepackage{epstopdf}
\usepackage{multirow}
\usepackage{array}
\usepackage{rotating}

\usepackage{url}

\usepackage{amssymb}

\usepackage{cancel}

\usepackage{array}
\newcolumntype{L}[1]{>{\raggedright\let\newline\\\arraybackslash\hspace{0pt}}m{#1}}
\newcolumntype{C}[1]{>{\centering\let\newline\\\arraybackslash\hspace{0pt}}m{#1}}
\newcolumntype{R}[1]{>{\raggedleft\let\newline\\\arraybackslash\hspace{0pt}}m{#1}}

\usepackage{tcolorbox}
\tcbuselibrary{theorems}

\definecolor{grey1}{RGB}{192,192,192}
\definecolor{grey2}{RGB}{178,178,178}
\definecolor{grey3}{RGB}{150,150,150}
\definecolor{grey4}{RGB}{119,119,119}
\definecolor{grey5}{RGB}{77,77,77}
\definecolor{green}{RGB}{109,169,69}
\definecolor{blue2}{RGB}{68,115,196}
\definecolor{red}{RGB}{192,0,0}
\definecolor{yellow}{RGB}{255,192,0}

\usepackage{cite}

\makeatletter
\def\namedlabel#1#2{\begingroup
    #2%
    \def\@currentlabel{#2}%
    \phantomsection\label{#1}\endgroup
}
\makeatother

\newcommand{\majorChange}{\textcolor{black}}
\newcommand{\majorChangeFinal}{\textcolor{black}}

\title{An Inversion-Based Learning Approach for Improving Impromptu Trajectory Tracking of Robots with Non-Minimum Phase Dynamics}

\author{Siqi Zhou$^{1}$, Mohamed K. Helwa$^{1,2}$, and Angela P. Schoellig$^{1}$%
\thanks{Manuscript received: September 10, 2017; Revised December 17, 2017; Accepted January 14, 2018.}
\thanks{This paper was recommended for publication by Editor Paolo Rocco upon evaluation of the Associate Editor and Reviewers' comments. 
This work was supported in part by OCE/SOSCIP TalentEdge Project \#27901 and various NSERC research and equipment grants.} 
\thanks{$^{1}$Siqi Zhou, Mohamed K. Helwa, and Angela P. Schoellig are with the  Dynamic  Systems  Lab  (www.dynsyslab.org), Institute for Aerospace Studies, University of Toronto, Canada. Emails: siqi.zhou@robotics.utias.utoronto.ca,  mohamed.helwa@robotics.utias.utoronto.ca,  schoellig@utias.utoronto.ca}%
\thanks{$^{2} $Mohamed K. Helwa is also with the Electrical Power and Machines Department, Cairo University, Egypt.}%
\thanks{Digital Object Identifier (DOI): see top of this page.}
}

\usepackage{fancyhdr}
\fancypagestyle{specialfooter}{%
\fancyhf{}
\fancyfoot[L]{Some test text}
\fancyfoot[C]{}
}
\newenvironment{copyrightnoticeFont}{\fontsize{7pt}{8pt}\selectfont\fontfamily{phv}\selectfont}{\par}

\fancypagestyle{frontPage}{%
    \fancyhead{}
    \cfoot{}
    \rhead{}
}

\begin{document}
\maketitle
\thispagestyle{frontPage}
\renewcommand{\headrulewidth}{0pt}
 \lfoot{\begin{copyrightnoticeFont}\vspace{-2.5em}
 \textbf{Accepted final version}. To appear in \textit{the IEEE Robotics and Automation Letters (RA-L).}\\
 \copyright2018 IEEE. Personal use of this material is permitted. Permission from IEEE must be obtained for all other uses, in any current or future media, including reprinting/republishing this material for advertising or promotional purposes, creating new collective works, for resale or redistribution to servers or lists, or reuse of any copyrighted component of this work in other works.\end{copyrightnoticeFont}}

\begin{abstract}
This paper presents a learning-based approach for impromptu trajectory tracking for non-minimum phase systems, i.e., systems with unstable inverse dynamics. Inversion-based feedforward approaches are commonly used for improving tracking performance; however, these approaches are not directly applicable to non-minimum phase systems due to their inherent instability. In order to resolve the instability issue, existing methods have assumed that the system model is known and used pre-actuation or inverse approximation techniques. In this work, we propose an approach for learning a stable, approximate inverse of a non-minimum phase baseline system directly from its input-output data. Through theoretical discussions, simulations, and experiments on two different platforms, we show the stability of our proposed approach and its effectiveness for high-accuracy, impromptu tracking. Our approach also shows that including more information in the training, as is commonly assumed to be useful, does not lead to better performance but may trigger instability and impact the effectiveness of the overall approach.
\end{abstract}

\begin{IEEEkeywords}
Model Learning for Control, Deep Learning in Robotics and Automation
\end{IEEEkeywords}





\section{Introduction}
\label{sec:introduction}





\IEEEPARstart{H}{igh-accuracy} trajectory tracking is essential for many robotic and automated systems. The concept of using the inverse dynamics to enforce high-accuracy or exact tracking is widely used in the control systems literature~\cite{clayton2009review}. However, for many practical problems ranging from aircraft control~\cite{al2002tracking} to flexible robot arm end-effector tracking~\cite{de1989inversion} and hard disk drive track-following~\cite{levin2009neural}, the input-output dynamics are non-minimum phase --- i.e., the inverse dynamics are inherently unstable. 
The non-minimum phase nature poses challenges in classical control design~\cite{hoagg2007nonminimum} and 
prohibits the direct application of inversion-based approaches.  \majorChange{Moreover, in this work, we consider the task of \textit{impromptu tracking} (i.e., tracking an arbitrary, feasible trajectory with high accuracy in one shot without further changing or tuning the control system)~\cite{DNNimpromptuTrack}, which is even more challenging to achieve.}

In the literature, 
various model-based inversion approaches have been proposed to resolve the instability issue associated with the system inverse of non-minimum phase systems. These approaches are based on \textit{(i)} pre-actuation~\cite{devasia1996nonlinear} or \textit{(ii)} inverse approximation~\cite{rigney2009nonminimum,slotine1991applied}. In the pre-actuation approach, first proposed in~\cite{devasia1996nonlinear}, a bounded input is ensured by pre-loading the system state to a desired initial state designed for the particular desired trajectory. Though exact tracking can be achieved with bounded input signals, the solutions are trajectory-specific and require significant setup time in order to reach the desired initial condition~\cite{zhang2016pre}. 
On the other hand, in the inverse approximation approaches, stability of the inverse is ensured by replacing the unstable components of the inverse dynamics with a stable approximation that is capable of achieving precise tracking (see~\cite{slotine1991applied,rigney2009nonminimum} and the references therein). 
As compared with the pre-actuation approaches, the approximate inversion approaches are more robust against modeling errors and consequent instability issues.
Moreover, since the inversion is system-specific, the approximate inversion approaches can be more easily generalized to impromptu tracking tasks. However, due to the model-based nature of both approaches, the effectiveness depends on 
sufficiently accurate system models. This limitation motivates the investigation of learning techniques, which leverage data to improve the performance of model-based approaches.


\begin{figure}[!t]
\centering
\vspace{0.5em}
\includegraphics[width=0.475\textwidth]{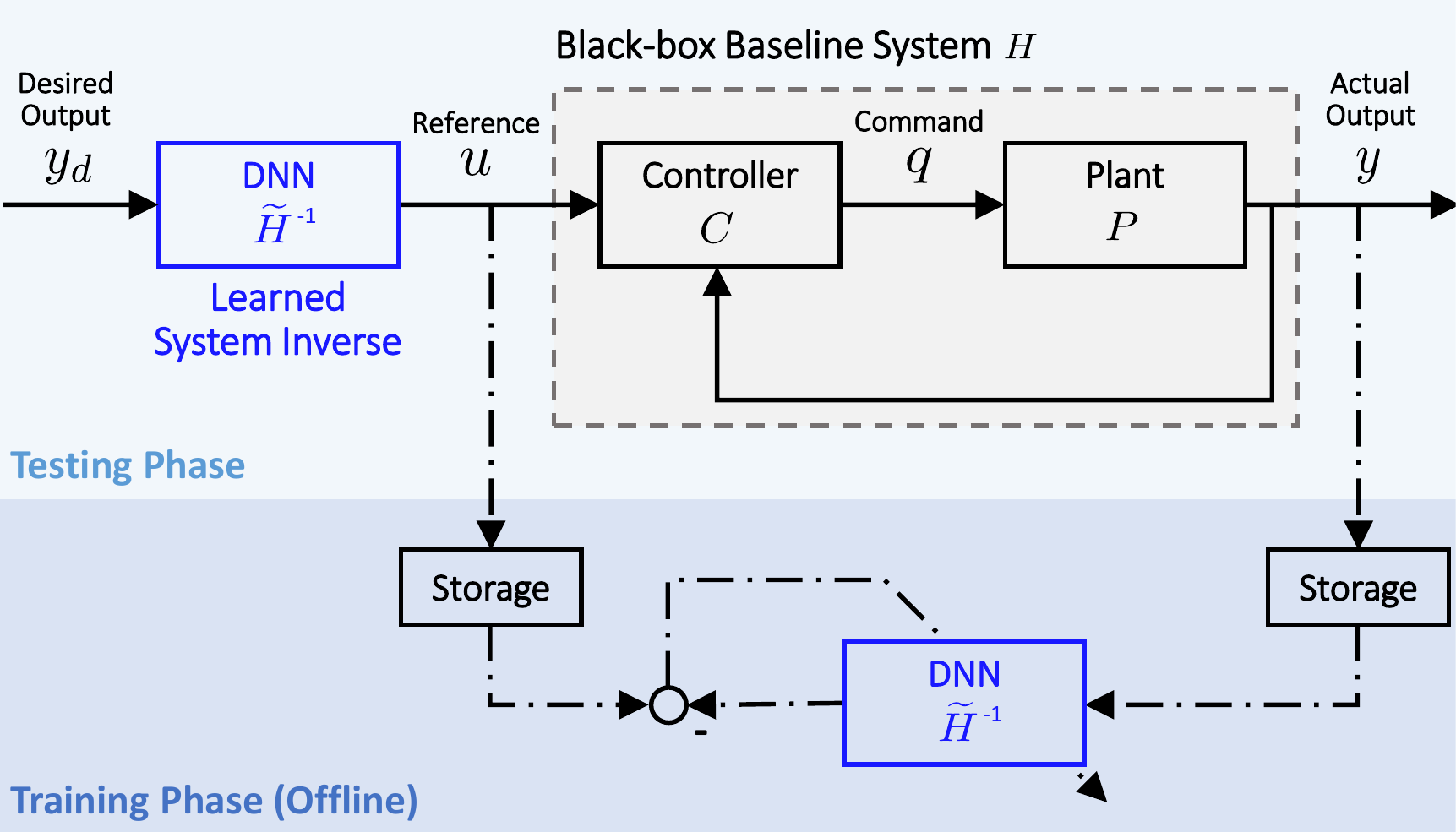}
\caption{An illustration of the proposed DNN-enhanced control architecture for output trajectory tracking. A stable baseline control system is treated as a black box and a DNN module is pre-cascaded to the baseline system to adjust reference signals to improve the tracking performance. 
}
\label{fig:blkdiag}
\vspace{-2em}
\end{figure}
For \textit{minimum phase systems}, different inverse dynamics learning approaches have been studied. In our previous work \cite{DNNimpromptuTrack,zhou-cdc17}, a deep-neural-network-based (DNN-based) architecture (Fig.~\ref{fig:blkdiag}) was proposed to enhance the tracking performance of minimum phase \majorChange{black-box systems (i.e., systems whose dynamical models are not available or not sufficiently accurate).} 
With experiments on quadrotors, it was shown that the proposed approach led to an average of 43\% tracking error reduction on 30 arbitrary, hand-drawn trajectories, as compared to the baseline controller. In addition to our previous work, the potential of utilizing inverse learning for high-accuracy tracking has been demonstrated using different robotic platforms and learning techniques (e.g., Gaussian Processes (GPs) and Locally Weighted Projection Regression (LWPR)), see for instance~\cite{polydoros2015real,nguyen2008learning,williams2009multi}. 
\majorChange{Nevertheless, the applicability of these inversion-based learning approaches to non-minimum phase systems has not been studied, and
systematically extending inverse dynamics learning schemes to non-minimum phase systems is still an open problem. }

Previously, for \textit{non-minimum phase systems}, a DNN-based adaptive feedback error learning approach has been proposed to learn an inverse of the open-loop plant for enhancing tracking~\cite{jung2008control,de2000feedback}. In this approach, the DNN training requires the plant or a good model of the plant in place, which may not always be desired in the initial training phase or available in practice. 
Moreover, similar to the adaptive inverse learning approaches discussed in~\cite{zhou-cdc17}, this approach is more susceptible to instability issues, especially when the DNN is not well-initialized~\cite{chen1995adaptive}.

\majorChange{In this paper, we present a learning-based approach that constructs an approximate inverse of a non-minimum phase, feedback-stabilized system based only on input-output data. 
In particular, informed by control theory, we select appropriate inputs and outputs of the inverse-learning module, prove stability of the learning-enhanced architecture for both linear and nonlinear systems, and provide theoretical insights on the inverse approximation utilized by the learning module to achieve performance enhancement. The efficacy of the proposed approach for nonlinear systems is verified with experiments on \emph{(i)} an inverted pendulum on a cart system, and \emph{(ii)} a modified non-minimum phase quadrotor system. For the quadrotor experiments, the generalizability of the learned inverse is verified by showing impromptu tracking of arbitrary, hand-drawn trajectories.
}
 \majorChange{Furthermore, we also show the connection between the proposed learning approach and a common model-based approximate inversion approach for linear systems~\cite{rigney2009nonminimum,slotine1991applied}. The proposed approach shares the same core concept as the model-based approach; yet, without requiring a detailed model, the proposed approach leads to better performance and is applicable to nonlinear systems.} 


\section{Problem Formulation}
\label{sec:problemFormulation}
\majorChange{We aim to provide an inversion-based learning approach for enhancing the tracking performance of non-minimum phase systems in impromptu tracking tasks.
} The proposed approach should satisfy the following objectives:
\begin{enumerate}
\item[\text{\namedlabel{objective:stability}{O1}}.] Stability --- the overall system, including the learning module, is input-to-output stable~\cite{sontag1999notions};
\item[\text{\namedlabel{objective:data}{O2}}.] Training --- the learning module relies only on the input-output data rather than a system model;
\item[\text{\namedlabel{objective:enhancement}{O3}}.] Performance and Generalizability --- with the learning module, the root-mean-square (RMS) tracking error is reduced for impromptu tracking tasks, compared to the baseline system.
\end{enumerate}

\subsection{Control Architecture}
We consider the inversion-based learning architecture shown in Fig.~\ref{fig:blkdiag}, which consists of a baseline system and a pre-cascaded, learned system inverse module enhancing the tracking performance via modifying the reference signal $u$. In the training phase, the input-output data, $u$ and $y$, generated from the baseline system is stored and used to construct a training dataset that typically has $y$ and $u$ at selected time steps as the labeled inputs and $u$ at the current time step as the labeled output. 
When later using the trained module in the testing phase, the desired trajectory $y_d$ is given to the learned inverse model as input (in place of $y$) to compute a reference $u$ that is sent to the baseline system. 

The considered architecture is different from typical inversion-based feedforward architectures where the inverse of the open-loop plant $P$ is used and the output signal from the inverse is directly applied to the plant~\cite{polydoros2015real,de2000feedback}. By learning the inverse of a stabilized baseline system, the proposed architecture decouples the performance enhancement problem  from 
the plant stabilization problem, which simplifies the design, analysis, and practical implementation~\cite{zhou-cdc17}. 







\subsection{System Representations}
We first motivate our proposed approach by analyzing linear time-invariant (LTI), single-input-single-output (SISO) systems and then extend our discussion to nonlinear SISO systems. 
For linear systems, we represent the baseline feedback control system by the transfer function
\begin{equation}
\label{eqn:transferFunction}
H(z) = \frac{Y(z)}{U(z)} = \frac{N(z)}{D(z)} = \frac{1+\sum_{i=1}^{n-r}\alpha_iz^i}{\sum_{i = 0}^n \beta_i z^{i}},
\end{equation}
where $U(z)$ and $Y(z)$ are the z-transforms of the input and output of the system, $N(z)$ and $D(z)$ are the numerator and denominator polynomials,
$n$ is the order of the system, $r$ is the relative degree of the system, and $\alpha_i,\beta_i\in\mathbb{R}$ are scalar constants. For nonlinear systems, we consider the control affine nonlinear system:
\begin{equation}
\label{eqn:stateSpace}
\begin{aligned}
x(k+1) = f(x(k)) + g(x(k))\:u(k),\:\:y(k) = h(x(k)),
\end{aligned}
\end{equation}
where $k\in \mathbb{Z}_{\ge 0}$ is the discrete time index, $x\in \mathbb{R}^{n}$ is the state, $u\in \mathbb{R}$ is the input, $y\in\mathbb{R}$ is the output, and $f(\cdot)$, $g(\cdot)$, $h(\cdot)$ are nonlinear smooth functions (i.e., functions for which all orders of differentiation exist and are continuous). 

\subsection{Assumptions}
In deriving a solution for our problem, we assume:
\begin{itemize}
\item[\text{\namedlabel{assumption:stability}{A1}}.] The underlying plant is stabilizable and the baseline system is stable;
\item[\text{\namedlabel{assumption:trajectory}{A2}}.] At any time instant $k$, the current and future values of the desired trajectory are known up to time $k+n$, where $n$ is the order of the baseline system;
\item[\text{\namedlabel{assumption:learning}{A3}}.] The learned inverse dynamics module are feedforward neural networks (FNNs) with (A3a) finite weights and biases and (A3b) continuous activation functions $\sigma (\cdot)$.
\end{itemize}
Assumptions \ref{assumption:stability} through \ref{assumption:learning} are reasonable in practice. \majorChange{For~\ref{assumption:stability}, well-developed control methods, including model-free controllers (e.g., PID controllers), can be used to stabilize a system even in the absence of a dynamical model.}
For \ref{assumption:trajectory}, a preview of $n$ time steps of the desired trajectory is typically available, and this assumption does not prevent combinations with on-line trajectory generation and adaptation algorithms. \majorChange{Moreover, for \ref{assumption:learning}, even though we use FNNs in this paper, the proposed approach can be potentially realized with other nonlinear regression techniques (e.g., GPs and LWPR). Assumption A3a can always be satisfied with standard DNN training algorithms, and assumption A3b holds for all common DNN activation functions (e.g., rectified linear units (ReLU), tanh, and sigmoid).}
\section{Non-Minimum Phase System Inverse Learning}
\label{sec:adaptationToNonMinimumPhase}
For non-minimum phase systems, one approach to resolve the instability issue in inversion-based approaches is to utilize stable inverse approximations. 
In this section, we adapt this concept to unknown, possibly nonlinear baseline systems using a 
DNN-based control architecture (Fig.~\ref{fig:blkdiag}). 
\subsection{Background on Exact Inverse Learning} %
\label{subsec:directApplication}

Given the control architecture in Fig.~\ref{fig:blkdiag}, in~\cite{zhou-cdc17}, it is shown that for a minimum phase system with a well-defined relative degree\footnote{See~\cite{zhou-cdc17} and the references therein for formal definitions of relative degree. The relative degree of a discrete-time system can be intuitively thought as the inherent time delay of the system. Experimentally, it is the number of time steps between the time at which an input is applied and the system first reacts.} $r$, exact tracking (i.e., $y(k+r)=y_d(k+r)$) can be achieved by training the DNN to model the \textit{exact inverse dynamics} of the baseline system. 
Following~\cite{zhou-cdc17}, for learning the exact inverse of system~\eqref{eqn:stateSpace}, the proper selection of inputs $\mathcal{I}$ and outputs $\mathcal{O}$ of the DNN module are $\mathcal{I} =\{x(k),y_d(k+r)\}$ and $\mathcal{O} = \{u(k)\}$. 
For LTI systems, based on the representation~\eqref{eqn:transferFunction},
the inputs of the DNN module can be  selected as 
\begin{equation}
\label{eqn:feature_tf}
\begin{aligned}
\mathcal{I} &=\{y_d(k\mathord{-}n\mathord{+}r:k\mathord{+}r),u(k\mathord{-}n\mathord{+}r:k\mathord{-}1)\},
\end{aligned}
\end{equation}
where 
consecutive time indices are abbreviated with `$:$' \cite{zhou-cdc17}. 
In practice, when applying these results to design the DNN module,
only basic system properties (i.e., $n$ and $r$) are needed. A system's order $n$ can be determined from basic physics laws, and the relative degree~$r$ can be determined from simple step-response experiments. Although the \textit{exact inverse learning} approach can be conveniently implemented in practice~\cite{DNNimpromptuTrack}, its effectiveness is restricted to \textit{minimum phase systems}~\cite{zhou-cdc17}. 
\subsection{The Proposed Approach: DNN Input Modification}
\label{subsec:proposedApproach}
We propose a learning approach that achieves stability~\eqref{objective:stability} and performance enhancement~\eqref{objective:enhancement} through modifying the DNN input selection. 
We first consider the linear baseline system~\eqref{eqn:transferFunction}, for which the exact inverse is
\begin{equation}
\label{eqn:exactInverse}
H^{-1}(z) = \frac{U(z)}{Y_d(z)}=\frac{D(z)}{N(z)} = \frac{\sum_{i = 0}^n \beta_i z^{i}}{1+\sum_{i=1}^{n-r}\alpha_iz^i},
\end{equation}
where $Y_d(z)$ is the $z$-transform of the desired output $y_d(k)$. 
For non-minimum phase systems, at least one root of the denominator $N(z)$ is outside of the unit circle, which is the source of instability that prevents the direct application of the inverse learning scheme in~\eqref{eqn:feature_tf}. If the input of the DNN module is selected such that the unstable dynamics associated with $N(z)$ cannot be learned, then the instability issues would not arise.
By applying the inverse $z$-transform to \eqref{eqn:exactInverse}, it can be shown that 
\begin{equation}
\label{eqn:exactInverseTimeSpace}
u(k)= \sum_{i=0}^n \beta_iy_d(k+i)  - \sum_{i=1}^{n-r} \alpha_iu(k+i),
\end{equation}
or
\begin{equation}
\label{eqn:stateSpaceDependencies}
u(k) = F(\underbrace{y_d(k:k\mathord{+}n)}_{\text{from $D(z)$}},\underbrace{u(k\mathord{+}1:k\mathord{+}n\mathord{-}r)}_{\text{from $N(z)$}}),
\end{equation}
where $F(\cdot)$ denotes a generic multi-variable function. From~\eqref{eqn:stateSpaceDependencies}, it can be seen that the unstable dynamics associated with $N(z)$ are reflected in the dependency of $u(k)$ on the sequence of reference signals $u(k\mathord{+}1:k\mathord{+}n\mathord{-}r)$. 

\vspace{0.5em}
\noindent\textbf{Proposed Input-Output Selection.} Based on~\eqref{eqn:stateSpaceDependencies}, we propose the following DNN input-output selection:
\begin{equation}
\label{eqn:feature_tf_nonminimumPhase}
\begin{aligned}
\mathcal{I} &=\{y_d(k:k\mathord{+}n)\}\text{ and }\mathcal{O} = \{u(k)\},
\end{aligned}
\end{equation}
where the sequence of $u$ is removed from the input $\mathcal{I}$ to prevent the DNN module from learning the unstable dynamics associated with $N(z)$.
\vspace{0.5em}

\majorChange{Note that, while the proposed input-output selection is derived based on linear systems, when applying the proposed approach to nonlinear systems, the DNN module learns an approximate inverse of the nonlinear baseline system rather than a linearized baseline system. This is due to the fact that the DNN module is directly trained with the input-output data generated by the nonlinear baseline system.}
\subsection{Stability of the Proposed Approach}
\label{subsec:stability}
The proposed approach was derived from~\eqref{eqn:transferFunction}
to guarantee stability for the LTI systems. In this subsection, we prove stability for nonlinear systems 
using assumptions \ref{assumption:stability} and \ref{assumption:learning}.


\vspace{0.5em}
\noindent\textbf{Lemma 1. Stability.} Consider the inversion-based learning control architecture in Fig.~\ref{fig:blkdiag} and the nonlinear system~\eqref{eqn:stateSpace}. Under assumptions \ref{assumption:stability} and \ref{assumption:learning}, 
the learning module input-output selection in~\eqref{eqn:feature_tf_nonminimumPhase} ensures that the overall control system (from $y_d$ to $y$) is input-to-output stable.
\vspace{0.5em}

\noindent\textit{Proof.}
From~\eqref{eqn:feature_tf_nonminimumPhase}, the learning module approximates a mapping from $\mathcal{I}=\{y_d(k\hspace{0.2em}\mathord{:}\hspace{0.2em}k\mathord{+}n)\}$ to $\mathcal{O}=\{u(k)\}$. 
For a typical $L$-layer FNN with $n+1$ inputs and 1 output, by denoting $\zeta_0(k) = [y_d(k)\; y_d(k+1)\: \cdots\: y_d(k+n) 
]^\intercal$ as the network input at time $k$, 
the output of a neuron $i$ in a hidden layer $l$, denoted by $\zeta_{l,i}(k)$, can be expressed as $\zeta_{l,i}(k) = \sigma \left(\sum_{j=1}^{N_{l-1}} w_{l,ij} \zeta_{l-1,j}(k) + b_{l,i}\right)$, 
where $\sigma(\cdot)$ is the activation function, $l \in\mathbb{N}$, $1\le l\le L-1$, is the layer index, $N_l\in \mathbb{N}$ is the number of neurons in layer~$l$, $\zeta_{l}\in\mathbb{R}^{N_l}$ is the output of the layer $l$, $w_l\in\mathbb{R}^{N_{l}\times N_{l-1}}$ and $b_l\in\mathbb{R}^{N_l}$ are the weights and bias associated with layer $l$, $\zeta_{l,i}$ and $\zeta_{l-1,j}$ are the $i$-th element of the vector $\zeta_{l}$ and the $j$-th element of the vector $\zeta_{l-1}$, $w_{l,ij}$ is the $i$-th row and $j$-th column element of the matrix $w_{l}$, and $b_{l,i}$ is the $i$-th element of the vector $b_{l}$. 
The output of the network is $\widetilde{F}(\zeta_0(k)) = \sum_{j=1}^{N_{L-1}} w_{L,1j}\zeta_{L-1,j}(k) + b_{L,1}$.
By assumptions~A3a and A3b, the network parameters $w$ and $b$ are bounded, and $\sigma$ is continuous; hence, the output of each neuron $i$ in layer $l$ (i.e., $\zeta_{l,i}$) is continuous in $\zeta_0$. Moreover, since $\widetilde{F}$ is a composition of $\zeta_{l,i}$, $\widetilde{F}$ is also continuous in $\zeta_0$.
Since every continuous function from a compact space into a metric space is bounded, the network output $u(k)$ is bounded for bounded input $\zeta_0(k)$. 
Furthermore, by assumption \ref{assumption:stability}, the baseline system is input-to-output stable; thus, for any bounded desired trajectory $y_d$, the output $u(k)$ of the FNN is bounded, and the overall system from $y_d$ to $y$ is input-to-output stable.~$\square$

\majorChange{Note that the input-to-output stability of the DNN module and the overall DNN-enhanced system rely on the fact that the proposed DNN module is a continuous, static mapping. This stability result holds for both linear and nonlinear systems and is independent of the DNN regression errors.} 
\subsection{Insights on Performance Enhancement}
\label{subsec:performanceEnhancement}

Given that the stability~\eqref{objective:stability} is achieved through the input selection of the learning module in~\eqref{eqn:feature_tf_nonminimumPhase}, 
in this subsection we address the performance enhancement objective~\eqref{objective:enhancement}. 
\vspace{0.5em}
\noindent\textbf{Insight 1. Approximate Inverse Learning.} For system~\eqref{eqn:transferFunction}, given a sufficiently high sampling rate, the input selection in~\eqref{eqn:feature_tf_nonminimumPhase} enables the FNN to learn an approximate inverse, where the sequence of reference signals in the input of the exact inverse map is approximated by $u(k)$.
\vspace{0.5em}

\begin{figure}[!t]
\vspace{1em}
\centering
    \includegraphics[trim={0 3.5em 0 3.5em},clip,width=0.42\textwidth]{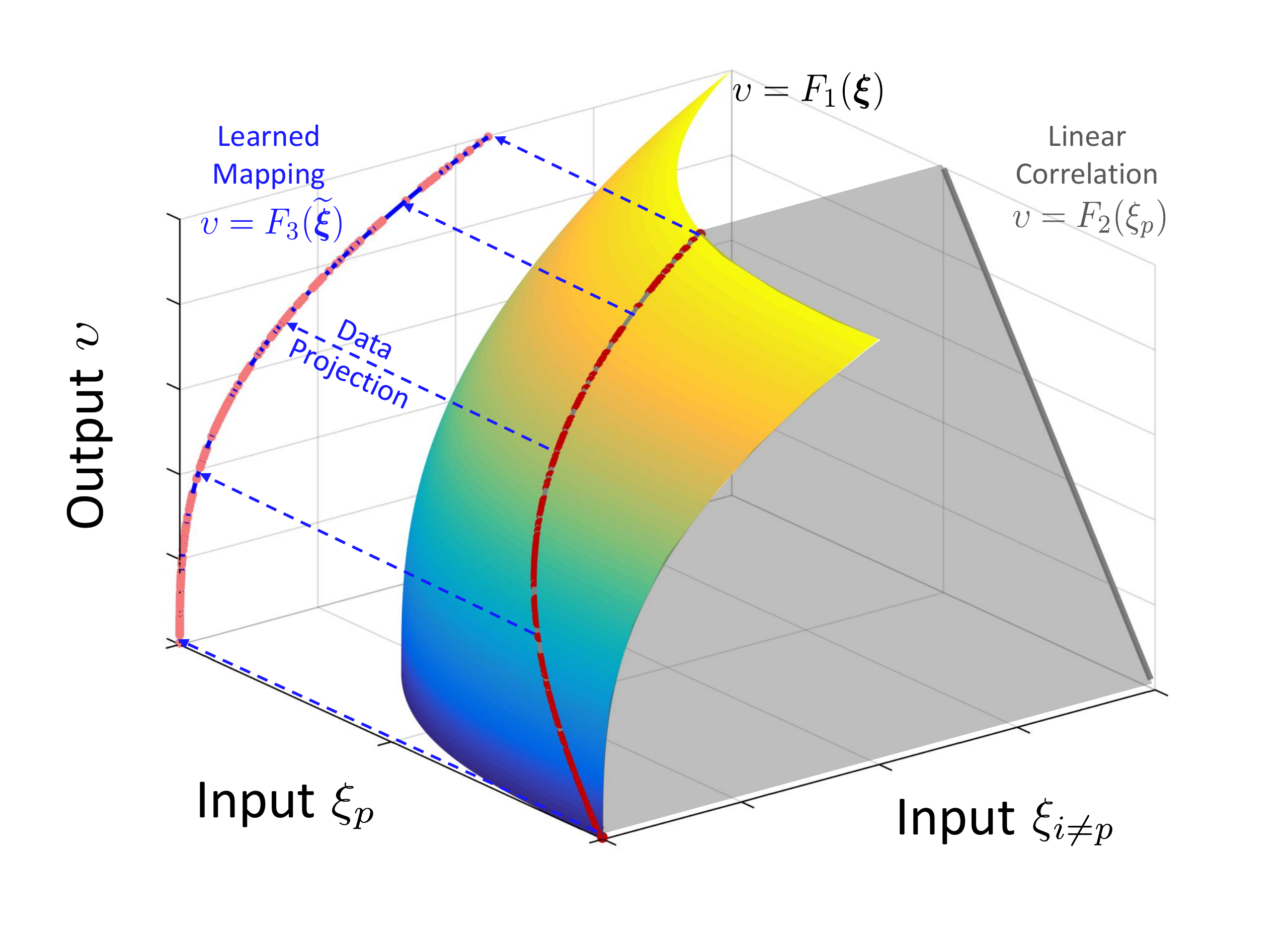}\\
\vspace{-1em}
    \caption{Illustration of data projection in the approximate inverse learning. 
}    \label{fig:inverseApproximationIllustration}
    \vspace{-1em}
\end{figure}

In order to clarify the insight above, we first present a toy example. Consider a linear function with input $\boldsymbol{\xi}= [\xi_1\;\xi_2\;...\;\xi_m]^\intercal\in\mathbb{R}^{m}$ and output $\upsilon \in \mathbb{R}$: $\upsilon = F_1(\boldsymbol{\xi})$. 
If a particular input $\xi_p$ is correlated to the output $\upsilon$ by the linear function $\upsilon = F_2(\xi_p)$ and $\frac{\partial F_1}{\partial \xi_p}\neq \frac{d F_2}{d \xi_p}$, then $\upsilon$ can be re-expressed as a linear function of the remaining components of the vector $\boldsymbol{\xi}$: $\upsilon = F_3(\widetilde{\boldsymbol{\xi}})$, 
where $\widetilde{\boldsymbol{\xi}}:=[
\xi_1 \; ... \; \xi_{p-1} \; \xi_{p+1}\;...\;\xi_m]^\intercal$. 
This implies that a regression model for the output $\upsilon$ can be found with either $\boldsymbol{\xi}$ or $\widetilde{\boldsymbol{\xi}}$ as the input. This simple discussion can be generalized to the case when the removal of the dimension $\xi_p$ \textit{does not} lead to a one-to-many map from $\widetilde{\boldsymbol{\xi}}$ to $\upsilon$; a regression model can be constructed in a lower-dimensional input space to uniquely determine the output $\upsilon$ for a given $\widetilde{\boldsymbol{\xi}}$. An illustration is shown in Fig.~\ref{fig:inverseApproximationIllustration}. 
When a component of the input vector is related to the output by the function $F_2$, the data points generated by $F_1$ are restricted to the intersection of the manifolds defined by $F_1$ and $F_2$. When $\xi_p$ is removed from the input of the dataset, the data points are projected onto a lower-dimensional space that is orthogonal to $\xi_p$. 

For training, since an arbitrary smooth trajectory can be expressed as a superposition of sinusoidal functions, without loss of generality, we consider in our discussion below a single sinusoidal training trajectory of the form $u(t)=A\sin(\frac{2\pi}{T} t)+b$, where $t$ denotes continuous time. 
It can be shown using Taylor series expansion of $u(t)$ that at time step $k$, future references $u(k+p)$ for $p=1,...,n-r$ can be related to the current reference $u(k)$ by
\begin{align}
\label{eqn:consecutiveReference}
u(k+p) 
= u(k) + \sum_{i=1}^\infty \left(\frac{2\pi p\Delta t}{T}\right)^{i} c_i(k),
\end{align}
where $\Delta t$ denotes the sampling time and $|c_i(k)|\le \frac{A}{i!}$. 
Given that $p$ is typically a small positive number bounded by $n-r$, if $\Delta t$ is sufficiently small as compared to the period of the trajectory $T$, then from~\eqref{eqn:consecutiveReference}, at a particular time step $k$, the future reference $u(k+p)$ and $u(k)$ are approximately correlated by the identity function. 
Given this approximate correlation and by the result above, though dependent reference components are removed from the FNN input based on the selection in \eqref{eqn:feature_tf_nonminimumPhase}, the FNN can still learn a regression model to output a reference $u$ that best matches that in the training dataset. Hence, the FNN acts as an approximate inverse from output $y$ to input $u$ that reduces the error between $y_d$ and $y$. From~\eqref{eqn:consecutiveReference}, the error involved in consecutive reference signal approximations and the inherent regression error in the learned inverse model is smaller for smaller $\Delta t$ (i.e., higher sampling frequency).

For nonlinear systems, to achieve exact tracking, the learning module should model the output equation of the inverse dynamics, and $u(k)$ should be a nonlinear function of $x(k)$ and $y_d(k+r)$ (see Section~\ref{subsec:directApplication}); however, for non-minimum phase systems, the internal instability of $x(k)$ can cause numerical issues~\cite{zhou-cdc17}. One trivial solution is to remove the state $x(k)$ from the DNN input and use $\mathcal{I}=\{y_d(k+r)\}$. Instead, we suggest to use the same proposed input selection as in~\eqref{eqn:feature_tf_nonminimumPhase}. A rough conjecture for this selection is as follows. Since smooth nonlinear systems can be approximated by piecewise affine/linear systems with arbitrary accuracy~\cite{helwa2015epsilon}, one can always represent the considered, smooth nonlinear system as an aggregation of local, $n$-dimensional, affine/linear models defined on local regions of a cover/partition of the nonlinear system state space. 
Since all models have order $n$, by following the derivation in Section~\ref{subsec:proposedApproach} for each local model, one obtains the same input selection as in~\eqref{eqn:feature_tf_nonminimumPhase} for each local model. Thus, it is reasonable to select the inputs for the DNN as in~\eqref{eqn:feature_tf_nonminimumPhase} even for nonlinear systems. The effectiveness of the proposed input selection for nonlinear systems is validated with simulations and experiments in Sections~\ref{sec:simulations} and \ref{sec:experiment}, respectively.
\subsection{Connection with the ZOS Approach}
\label{subsec:connectionZOS}
In this subsection, we show a connection between the proposed approach and a model-based approximate inverse approach for linear systems, the zero-order series (ZOS) approach~\cite{rigney2009nonminimum}. 
In the ZOS approach, the transfer function polynomials associated with the unstable zeros are approximated by zero-order Taylor series~\cite{rigney2009nonminimum}. In particular, by re-expressing~\eqref{eqn:transferFunction} as $H(z) = \frac{N_s(z)N_u(z)}{D(z)}$, 
 the ZOS approximate inverse is
\begin{align}
\label{eqn:zosApproximation}
\widetilde{H}^{-1}_{\text{ZOS}}(z) = \frac{D(z)}{N_u(z)\vert_{z=1}N_s(z)},
\end{align}
where $N_s(z)$ and $N_u(z)$ denote the numerator polynomials with stable and unstable zeros, respectively.

\vspace{0.5em}
\noindent\textbf{Insight 2. Connection with ZOS.} For linear systems, the approximation of the sequence of reference signals with the current reference $u(k)$ is equivalent to approximating the numerator of the transfer function $N(z)$ in~\eqref{eqn:transferFunction} with $N(z)\vert_{z=1}$. With the input selection in~\eqref{eqn:feature_tf_nonminimumPhase}, the proposed learning approach achieves stability (\ref{objective:stability}) and performance enhancement (\ref{objective:enhancement}) in a similar manner as the model-based ZOS approach in~\eqref{eqn:zosApproximation}.
\vspace{0.5em}

The time-domain representation of the exact inverse in~\eqref{eqn:exactInverse} is shown in~\eqref{eqn:exactInverseTimeSpace}. When $u(k+i)$ for $i \mathord{=} 1,...,n\mathord{-}r$ are approximated by $u(k)$ as in the proposed approach, we obtain $\sum_{i=0}^n \beta_iy(k+i) \approx \left(1 + \sum_{i=1}^{n-r} \alpha_i\right)u(k)$, or
$H^{-1}(z) \approx \frac{\sum_{i=0}^n \beta_i z^i}{1 + \sum_{i=1}^{n-r} \alpha_i}=\frac{D(z)}{N(z)\vert_{z=1}}$ in the $z$-domain.
By comparing the latter expression with the ZOS approximation in~\eqref{eqn:zosApproximation}, it can be seen that 
they both achieve stability by approximating unstable zero dynamics at $z=1$, and compensating for the delays introduced by the dynamics associated with the poles ($D(z)$) to improve tracking performance. 

\majorChange{Note that the generalizability of the FNN depends on the invariance of the phase and magnitude errors of the transfer function $\frac{Y(z)}{Y_d(z)}=\frac{N(z)}{N(z)\vert_{z=1}}$ with respect to the frequency of the desired trajectory; it can be shown that the generalizability is better if the zeros (the roots of $N(z)$) are further away from $z = 1$.} \majorChange{Moreover, similar to the ZOS approach~\cite{slotine1991applied}, we expect that the proposed learning approach is more effective for enhancing the tracking performance of desired trajectories with frequencies less than the frequency of the zeros.} 


\section{Simulation Results}
\label{sec:simulations}
We use an inverted pendulum on a cart system (pendulum-cart system) to illustrate the efficacy of the proposed approach for nonlinear non-minimum phase systems. 



\subsection{Simulation Setup}
\label{subsec:pendulum_setup}

The pendulum-cart system has two degrees of freedom -- the cart linear position~$\eta$ and the pendulum angular position~$\theta$. 
By applying Lagrangian's equations, a dynamics model of the pendulum-cart system can be obtained \cite{bloch2000controlled}:
\begin{equation}
\label{eqn:nonlinearEQM}
\begin{aligned}
\ddot{\eta}&=\frac{q+mg\sin\theta\cos\theta-ml\dot{\theta}^2\sin\theta}{M+m\sin^2\theta}\\
\ddot{\theta}&=\frac{q\cos\theta+(M+m)g\sin\theta-ml\dot{\theta}^2\sin\theta\cos\theta}{l\left(M+m\sin^2\theta \right)},
\end{aligned}
\end{equation}
where $M$ and $m$ are the masses of the cart and the pendulum, respectively, $l$ is the effective length of the pendulum relative to the pivot point, and $q$ is the force applied to the cart. 
By defining the state of the system as $x = [ \eta \; \dot{\eta} \; \theta \; \dot{\theta} ]^\intercal$, its input as the force $q$, and its output as the full state $y=x$, the nonlinear state-space representation of the pendulum-cart system can be written in the control affine form:
\begin{equation}
\label{eqn:state_space_pendulum}
\begin{aligned}
\dot{x} &= f_1(x_2,x_3,x_4)+g_1(x_3)\:q,\:\:
y=x,
\end{aligned}
\end{equation}
where $x_2=\dot{\eta}$, $x_3=\theta$, and $x_4=\dot{\theta}$.
The control objective is to compute a control input $q$ such that the cart tracks a desired trajectory $\eta_d(t)$ while the pendulum is balanced at the upright position. The desired output is $y_d(t)=[
\eta_d(t)\hspace{1em}\dot{\eta}_d(t)\hspace{1em} 0 \hspace{1em} 0
]^\intercal$. \majorChange{Through linearizing the system~\eqref{eqn:state_space_pendulum} at $\eta=\eta_d$, $\dot{\eta}=0$, $\theta = 0$, $\dot\theta = 0$, and $q=0$, the pole placement technique can be used to find a stabilizing controller $q(t) = K_1(u(t) - y(t))$, where $u$ is the reference of the baseline system and for our simulations $K_1 = [\mathord{-}0.8678 \:\:  \mathord{-}1.808   \:\: 25.46   \:\:  4.140 ]$.}

A learning module, pre-cascaded to the baseline system as in Fig.~\ref{fig:blkdiag}, is designed based on~\eqref{eqn:feature_tf_nonminimumPhase} to enhance the performance of the cart position tracking. Given the desired trajectory $\eta_d$ (a component of $y_d$), at a time instance $k$, the learning module computes an adjusted reference signal $\eta_r$ (a component of $u$) to be sent to the baseline system. The $\dot{\eta}_r$ component in $u$ is generated from the $\eta_r$ trajectory.  
An FNN with 2 hidden layers of 5 hyperbolic tangent neurons is used for learning the approximate inverse of the baseline system. Assuming that the baseline system succeeds to stabilize the pendulum at the upright position, then from~\eqref{eqn:nonlinearEQM}, the dynamics associated with $\eta$ may be approximated by a second-order system; by~\eqref{eqn:feature_tf_nonminimumPhase}, 
the input and output of the learning module are selected to be $\mathcal{I}=\{\eta_d(k\hspace{0.2em}\mathord{:}\hspace{0.2em}k\mathord{+}2)\}$ and $\mathcal{O}=\{\eta_r (k)\}$. The learning module is executed at sampling intervals of 0.015~s. \majorChange{The module is trained on 30 sinusoidal trajectories with different combinations of amplitudes $\{0.5, 1.0, 1.5, 2.0, 2.5, 3.0\}$~m and periods $\{5, 10, 15, 20, 25\}$~s.} \majorChangeFinal{The training dataset consists of pairs of $(\mathcal{I}=\{\eta(k\hspace{0.2em}\mathord{:}\hspace{0.2em}k\mathord{+}2)\}, \mathcal{O}=\{\eta_r (k)\})$ randomly sampled from the 30 training trajectories with equal proportions. Validation of the FNN model is performed on 30\% of the training dataset; additional validation of the learning module is done by running the overall system on untrained trajectories.} 

\subsection{Results}
\begin{figure}[!t]
\centering
\vspace{1em}
\includegraphics[trim={0 0 0 0},clip,width=0.475\textwidth]{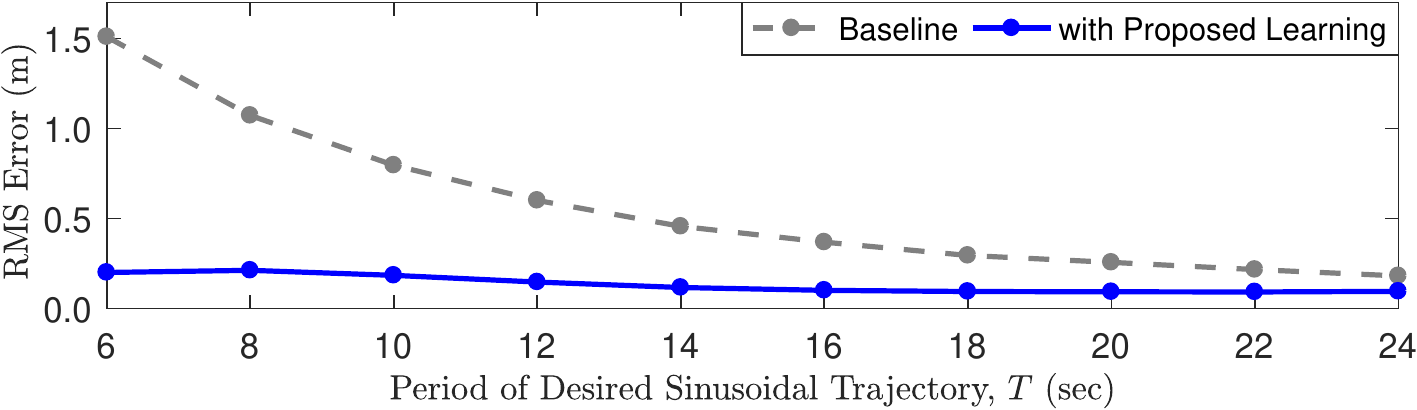}
\vspace{-0.8em}
\caption{The RMS tracking error of the baseline and the learning-enhanced system for desired trajectories of the form $\eta_d(t)=\frac{5}{2}\sin\left(\frac{2\pi}{T}t\right)$, where the periods $T$ are different from those used for training. The RMS error reduction achieved by the learning module ranges from 47\% to 87\%. 
A video for $T = 12$~s can be found at: \href{http://tiny.cc/fq0mny}{http://tiny.cc/fq0mny}.
\vspace{-0.5em}
}
\label{fig:simulation_rms_summary}
\end{figure}

\majorChange{The tracking performance of the baseline system and the learning-enhanced system are compared in Fig.~\ref{fig:simulation_rms_summary} for test sinusoidal trajectories with frequencies different from those used in training.} 
From Fig.~\ref{fig:simulation_rms_summary}, although the baseline system is capable of stabilizing the pendulum-cart system, the tracking error increases with decreasing periods of desired trajectories. In contrast, when the proposed learning module is added to the baseline system, the tracking error is approximately maintained at a smaller constant value over the range of trajectory periods covered by the training dataset, which shows the generalizing capabilities of the learning approach. 

\begin{figure}[!t]
\centering
\includegraphics[trim={0 0.01cm 0 0},clip,width=0.475\textwidth]{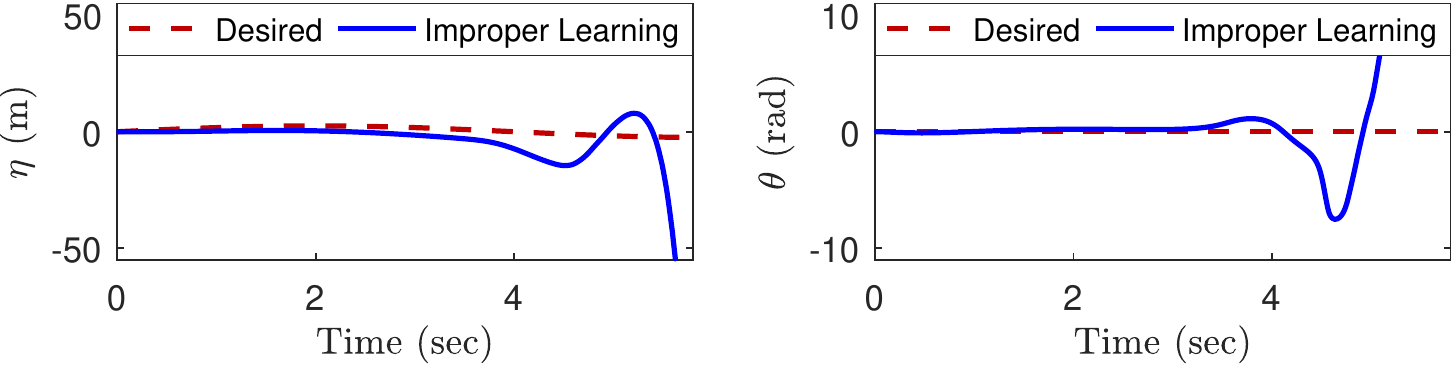}
\vspace{-0.8em}
\caption{Illustration of the adverse effect caused by the inclusion of an additional reference component in the input $\mathcal{I}$ of the learning module.}
\label{fig:simulation_improperSelection}
\vspace{-1.2em}
\end{figure}

Fig.~\ref{fig:simulation_improperSelection} shows the adverse impact when a single past reference is included in the proposed input selection of the learning module, i.e., when $\mathcal{I}=\{\eta_d(k\hspace{0.2em}\mathord{:}\hspace{0.2em}k\mathord{+}2),u(k\mathord{-}1)\}$. It can be seen that 
when the additional information is included,  
the pendulum-cart system quickly becomes unstable. Thus, for non-minimum phase systems, the input selection of the learning module is essential; the inclusion of unnecessary inputs can prevent not only the learning approach but also the baseline system from being functional. From this example, it is interesting to see that, for non-minimum phase systems, \textit{the DNN trained with less inputs leads to a better performance}. 
In contrast to typical DNN applications (e.g., image classification), for control applications, the training objective (e.g., minimizing regression error) and performance objective (e.g., minimizing tracking error) may not coincide. Consequently, DNN training algorithms may not phase out unnecessary input dimensions to achieve a good performance.
\section{Experimental Results}
\label{sec:experiment}
The effectiveness of the proposed approach is further verified using pendulum-cart and quadrotor experiments. \majorChange{Note that, in the experiments, the criterion we use for evaluating tracking performance is the RMS tracking error, which characterizes tracking performance over entire trajectories.}
\subsection{Pendulum-Cart Experiments}
\subsubsection{Experiment Setup}
The setup is similar to that of the simulation (Section~\ref{subsec:pendulum_setup}), except that the input force $q$ is replaced by the input voltage $v$ to the cart motor. 
By using a simple voltage-to-force model $q(t) = -7.74\dot\eta(t) + 1.73v(t)$~\cite{quadsor}, system~\eqref{eqn:state_space_pendulum} can be re-expressed as
\begin{equation}
\label{eqn:state_space_pendulum_experiment}
\begin{aligned}
\dot{x} = f_2(x_2,x_3,x_4)+g_2(x_3 )\:v,\:\:
y&=x,
\end{aligned}
\end{equation}
where $\eta(t)$ and $\theta(t)$ are measured, and
$x(t)$ is estimated with 
a full-state observer. A controller $v(t)=K_2(u(t)-y(t))$ with $K_2=[\mathord{-}105.6\:\: \mathord{-}55.04\:\: 130.7\:\: 23.67]$ is run at 1~kHz.

We compare the proposed learning approach with the baseline system and the model-based ZOS approach. 
In the experiments, the learning module is run at 70 Hz~\cite{DNNimpromptuTrack}; the design and training procedure for the inverse-learning module are similar to that of the simulations (see Section~\ref{sec:simulations}). 
\majorChange{The training dataset is constructed from 18 sinusoidal trajectories with combinations of amplitudes $\{0.04,0.06,0.08\}$~m and periods $\{5,6,7,8,9,10\}$~s.} 
The ZOS approach is implemented based on the linearized state-space model of system~\eqref{eqn:state_space_pendulum_experiment}. From the linearized system, a discrete-time transfer function from the reference $\eta_r$ to the output $\eta$ can be determined. By applying~\eqref{eqn:zosApproximation}, the ZOS approximate inverse is obtained: $\widetilde{H}_{\text{ZOS}}^{-1}(z)=\frac{ z^4 - 3.5217 z^3 + 4.6504 z^2 - 2.7290 z + 0.6005}{0.00137 z^2 - 0.0001066 z - 0.001066}$.
For the experimental comparison, the ZOS approximate inverse $\widetilde{H}_{\text{ZOS}}^{-1}(z)$ replaces the learning module in Fig.~\ref{fig:blkdiag}.

\subsubsection{Results}
\begin{figure}[!t]
\centering
\vspace{1em}
\includegraphics[trim={0 0.03cm 0 0},clip,width=0.475\textwidth]{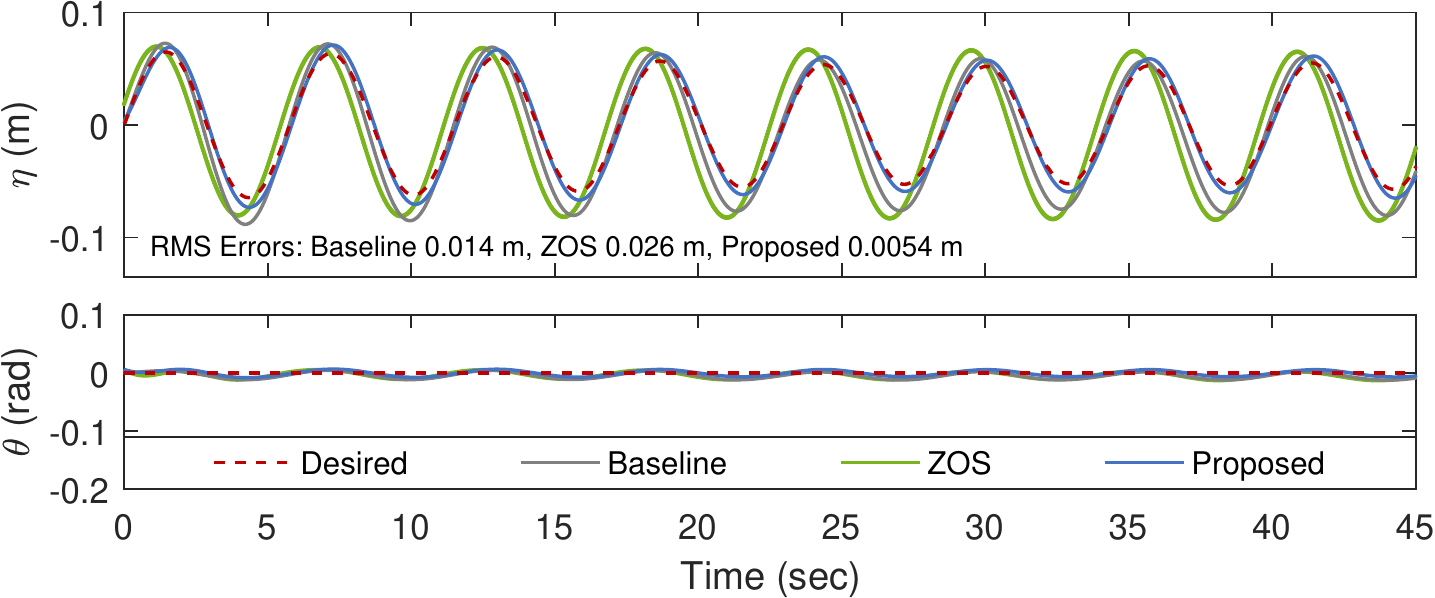}
\vspace{-0.8em}
\caption{The cart position $\eta$ and the pendulum angle $\theta$ of the baseline, the ZOS, and the proposed learning-based systems on a test trajectory $\eta_d(t) = \frac{117}{2000}\sin(\frac{2\pi}{5} t)+\frac{13}{2000}\sin(\frac{4\pi}{11} t)$.}
\label{fig:experimentResult}
\vspace{-1.5em}
\end{figure}
\majorChange{Fig.~\ref{fig:experimentResult} shows the comparison of the tracking performance of the baseline, the ZOS, and the proposed learning-based systems on a test trajectory $\eta_d(t) = \frac{117}{2000}\sin(\frac{2\pi}{5} t)+\frac{13}{2000}\sin(\frac{4\pi}{11} t)$, which was not included in the training phase.} 
The stability objective is achieved by all three systems, and the pendulum position is kept approximately at the upright position. From the cart position $\eta(t)$ plot, the proposed DNN (blue) effectively compensates for the phase and magnitude errors in the baseline system response (gray). 
For this test trajectory, the learning module reduces the RMS tracking error by 60\%. 

In contrast, by comparing the $\eta(t)$ of the ZOS approach (green) with the baseline response (gray), the addition of the approximate inversion led to worse tracking performance. Though the linearized state-space model is sufficiently accurate for deriving a baseline controller that stabilizes the pendulum-cart system, the application of the model-based system inversion approach requires a much more detailed and accurate system model.  
Thus, in comparison with the ZOS approach, \textit{the proposed DNN-based learning approach (blue) is capable of achieving a better performance without relying on a detailed dynamic model of the baseline system. 
}




\color{black}
\subsection{Quadrotor Experiments}
\begin{figure}[!t]
\centering
\vspace{1em}
\includegraphics[trim={0cm 0.1cm 0cm 0cm},clip,width=0.465\textwidth]{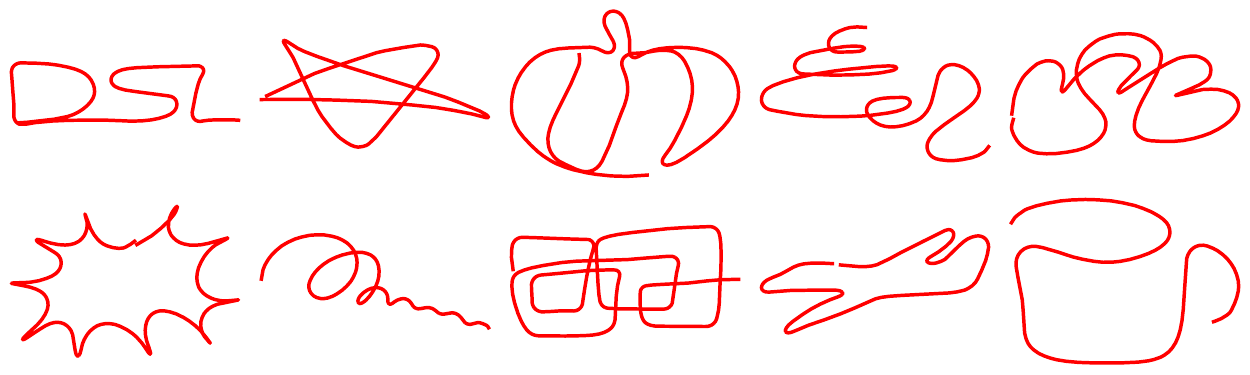}
\vspace{-0.5em}
\caption{\majorChange{Illustrations of 10 hand-drawn test trajectories used for evaluating the tracking performance of the quadrotor controllers.}}
\label{fig:testTrajectories}
\vspace{-1.5em}
\end{figure}

The efficacy of the proposed approach on higher degree-of-freedom systems is demonstrated using quadrotor vehicles. 
In this set of experiments, the objective is to enhance a baseline controller of a quadrotor for tracking arbitrary, hand-drawn trajectories (Fig.~\ref{fig:testTrajectories}) in one shot~\cite{DNNimpromptuTrack,zhou-cdc17}.

\subsubsection{Experiment Setup}
The state vector of the quadrotor system consists of the positions $\mathbf{p}=(x,y,z)$, velocities $\mathbf{v} = (\dot{x},\dot{y},\dot{z})$, roll-pitch-yaw Euler angles $\boldsymbol{\theta} = (\phi,\theta,\psi)$, and rotational velocities $\boldsymbol{\omega} = (p,q,r)$. The control objective is to control the position of the quadrotor to track a desired trajectory $\mathbf{p}_d(t)$. The baseline tracking controller is a standard nonlinear controller composed of a nonlinear transformation and PD control~\cite{DNNimpromptuTrack} running at 70~Hz. For the purpose of studying non-minimum phase systems, non-minimum phase zeros at $1.2$ are introduced to the baseline system by modifying the baseline $z$ position and velocity references ($z_r$ and $\dot{z}_r$). Note that, in this paper, we purposely introduce a non-minimum phase zero to the baseline system for evaluating our proposed approach; in practice, this non-minimum phase nature can occur in apparent minimum phase robotic systems when the sampling rate is high~\cite{butterworth2008effect}. 

\begin{figure}[!t]
\centering
\includegraphics[width=0.475\textwidth]{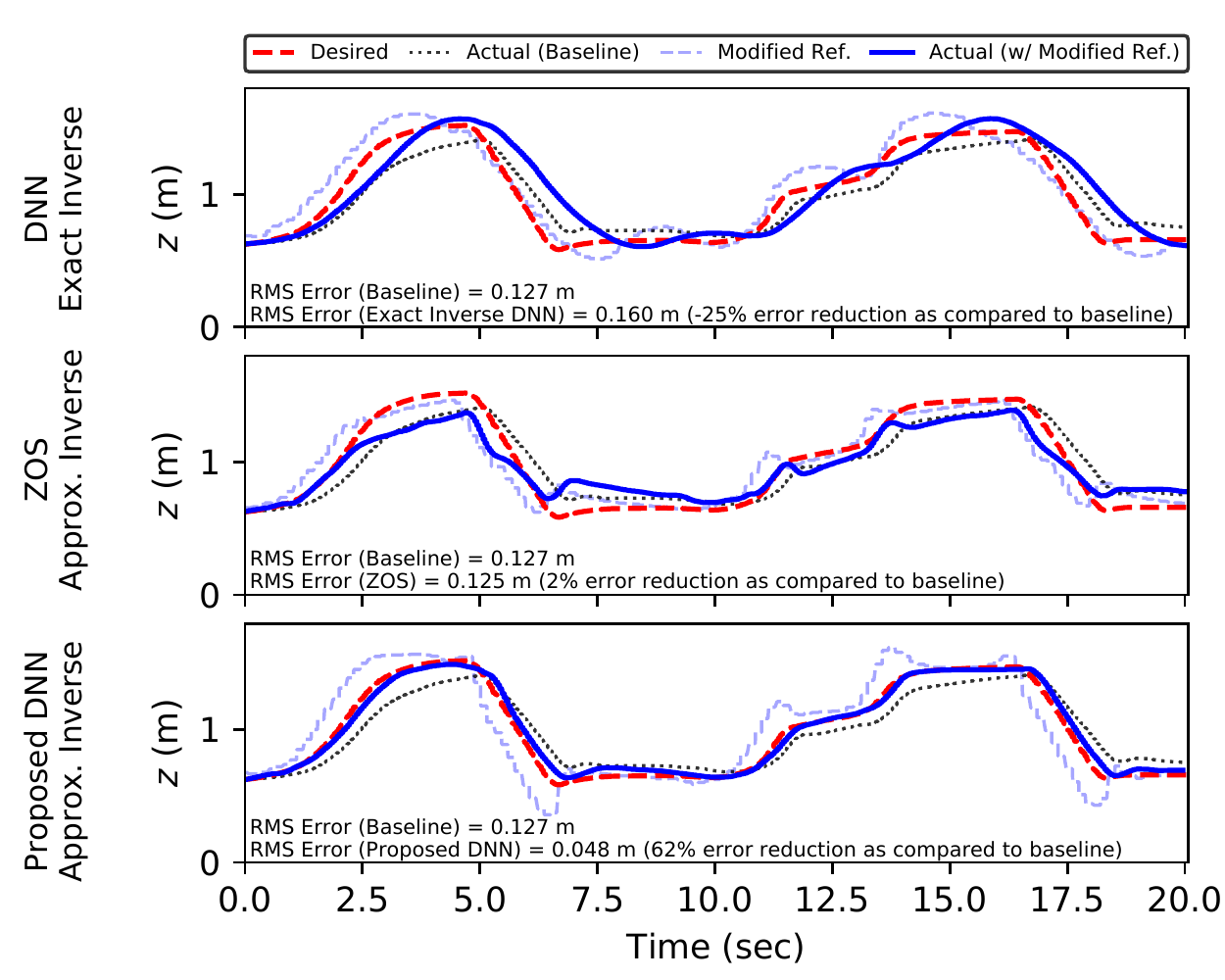}
\vspace{-1em}
\caption{\majorChange{Comparison of the DNN exact inverse approach~\cite{DNNimpromptuTrack,zhou-cdc17}, the ZOS approximate inverse approach, and the proposed DNN approximate inverse approach for enhancing the tracking performance of the modified non-minimum phase quadrotor system. The desired $z$-position trajectory is from the first hand-drawn test trajectory shown in Fig.~\ref{fig:testTrajectories} (left, top). 
}}
\label{fig:z_trajectory}
\vspace{-0.8em}
\end{figure}
In the experiments, we examine three inversion-based approaches that adapt the reference signals of the baseline controller $\mathbf{p}_r$ and $\mathbf{v}_r$ to reduce the tracking error between the desired position $\mathbf{p}_d$ and the actual position $\mathbf{p}$:
\begin{enumerate}[leftmargin=2.8\parindent]
\item[(M1)] DNN exact inverse learning: the learning-based approach effective for minimum phase systems~\cite{DNNimpromptuTrack,zhou-cdc17};
\item[(M2)] ZOS approximate inverse: a model-based approach for non-minimum phase systems;
\item[(M3)] DNN approximate inverse learning: the proposed learning-based approach with input-output selection based on~\eqref{eqn:feature_tf_nonminimumPhase}.
\end{enumerate}
The inverse blocks receive the desired position $\mathbf{p}_d$ and desired velocity $\mathbf{v}_d$ as inputs, and compute the adjusted position reference $\mathbf{p}_r$ and velocity reference $\mathbf{v}_r$ for the baseline system. 
For comparison purposes, the DNN training and architecture are similar to~\cite{DNNimpromptuTrack,zhou-cdc17}. In particular, the DNNs are fully-connected feedforward networks with 4 hidden layers of 128 ReLUs. 
\majorChangeFinal{During the training phase, the baseline system is used to track a 400-second, 3-dimensional sinusoidal trajectory, and the input-output data of the baseline system is collected at 7~Hz. The training dataset of the DNN consists of $\left(\mathcal{I},\mathcal{O}\right)$ pairs randomly sampled from the input-output data of the baseline system. Overall, 90\% of the dataset is used for training, and the remainder of the dataset is used for validation.} 
For evaluating the effectiveness and generalizability of the inversion-based approaches, test trajectories generated from arbitrary hand drawings are utilized (Fig.~\ref{fig:testTrajectories}).













\subsubsection{Results}
We first examine the three inversion-based approaches for enhancing the tracking performance of the modified non-minimum phase quadrotor baseline system, where non-minimum phase zeros are introduced in the dynamics associated with the $z$-direction. The implementation of (M1) follows from that in \cite{zhou-cdc17}; the inputs and outputs of the DNN are selected to be $\mathcal{I}=\{x_d(k+5)-x(k),y_d(k+5)-y(k),z_d(k+3)-z(k),\dot{x}_d(k+4)-\dot{x}(k),\dot{y}_d(k+4)-\dot{y}(k),\dot{z}_d(k+2)-\dot{z}(k),\boldsymbol{\theta}(k),\boldsymbol{\omega}(k)\}$ and $\mathcal{O}=\{x_r(k)-x(k),y_r(k)-y(k),z_r(k)-z(k),\dot{x}_r(k)-\dot{x}(k),\dot{y}_r(k)-\dot{y}(k),\dot{z}_r(k)-\dot{z}(k)\}$. The implementation of (M2) is based on the approximation of the dynamics of the baseline system with decoupled second-order linear systems; by applying Eqn.~\eqref{eqn:zosApproximation}, the ZOS approximate inverse is found to be $H^{-1}_\text{ZOS} (z)= \frac{z^3 - 1.713 z^2 + 0.7493 z}{ 0.2692 z - 0.2331}$, and is applied to adjust the position and velocity references $z_r$ and $\dot{z}_r$. In the implementation of (M3), we need to estimate the system order $n$. We assume that the quadrotor has decoupled double-integrator dynamics in the $x$, $y$, and $z$ directions. By further accounting for the experimentally determined time delays in each direction and applying~\eqref{eqn:feature_tf_nonminimumPhase}, the inputs and outputs of the DNN module are selected to be $\mathcal{I} =\{x_d(k\mathord{+}1:k\mathord{+}7)-x_d(k), y_d(k\mathord{+}1:k\mathord{+}7)-y_d(k),z_d(k\mathord{+}1:k\mathord{+}5)-z_d(k),\dot{x}_d(k\mathord{+}1:k\mathord{+}6)-\dot{x}_d(k), \dot{y}_d(k\mathord{+}1:k\mathord{+}6)-\dot{y}_d(k),\dot{z}_d(k\mathord{+}1:k\mathord{+}4)-\dot{z}_d(k)\}$ and  $\mathcal{O}=\{x_r(k)-x_d(k),y_r(k)-y_d(k),z_r(k)-z_d(k),\dot{x}_r(k)-\dot{x}_d(k),\dot{y}_r(k)-\dot{y}_d(k),\dot{z}_r(k)-\dot{z}_d(k)\}$. Following previous work~\cite{DNNimpromptuTrack,zhou-cdc17}, in the implementations of (M1) and (M3), we utilized a difference learning scheme (i.e., training with relative positions and velocities) 
to improve training efficiency. 

\begin{figure}[!t]
\centering
\includegraphics[width=0.45\textwidth]{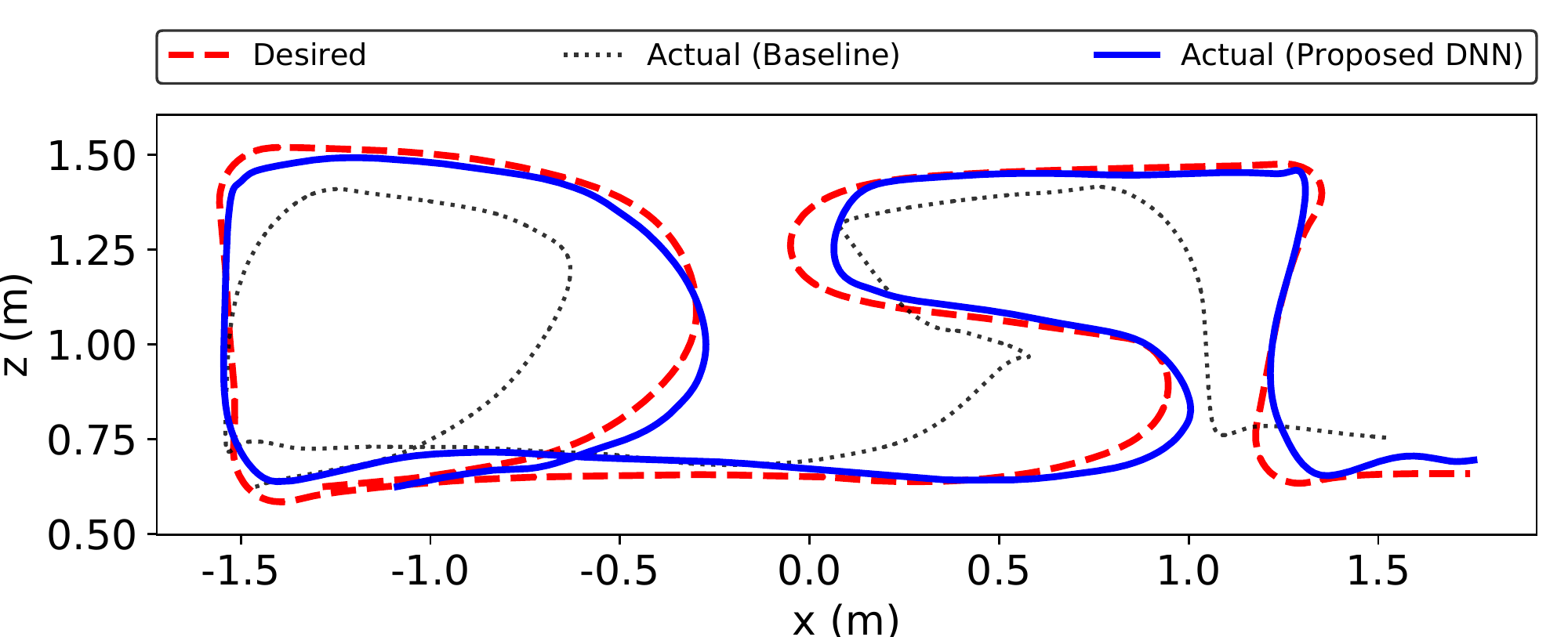}
\vspace{-0.8em}
\caption{\majorChange{Example of the performance enhancement achieved by the proposed DNN approximate inverse approach for the modified non-minimum phase quadrotor system. Here, the proposed DNN leads to a 67\% error reduction.}
}
\label{fig:dsl_tracking}
\vspace{-1.8em}
\end{figure}

Fig.~\ref{fig:z_trajectory} shows a comparison of the three inversion-based approaches for a test trajectory in the $z$-direction, $z_d(t)$, from the first hand drawing shown in Fig.~\ref{fig:testTrajectories}. 
From the top panel, as expected, due to the inherent instability of the inverse, the approach (M1) does not lead to an improved tracking performance. Instead, it introduces undesired oscillations in the system response and leads to worse performance as compared with the baseline controller. We next consider (M2) shown in the middle panel. From the computed reference $z_r$ (light blue dotted line), it can be seen that the model-based system approximate inversion tends to compensate for the delay in the system response; however, with the linearized model, the approximate inverse $H^{-1}_\text{ZOS}$  cannot effectively reduce the magnitude error of the system response. In contrast, for the proposed approach (M3), shown in the bottom panel, the reference computed by the DNN module efficaciously compensates for the tracking errors of the baseline response. With (M3), the RMS tracking error in the $z$-direction is reduced by approximately 62\%, while the percentage reductions for (M1) and (M2) are approximately -25\% and 2\%, respectively.

\begin{figure}[!t]
\centering
\vspace{1em}
\includegraphics[width=0.475\textwidth]{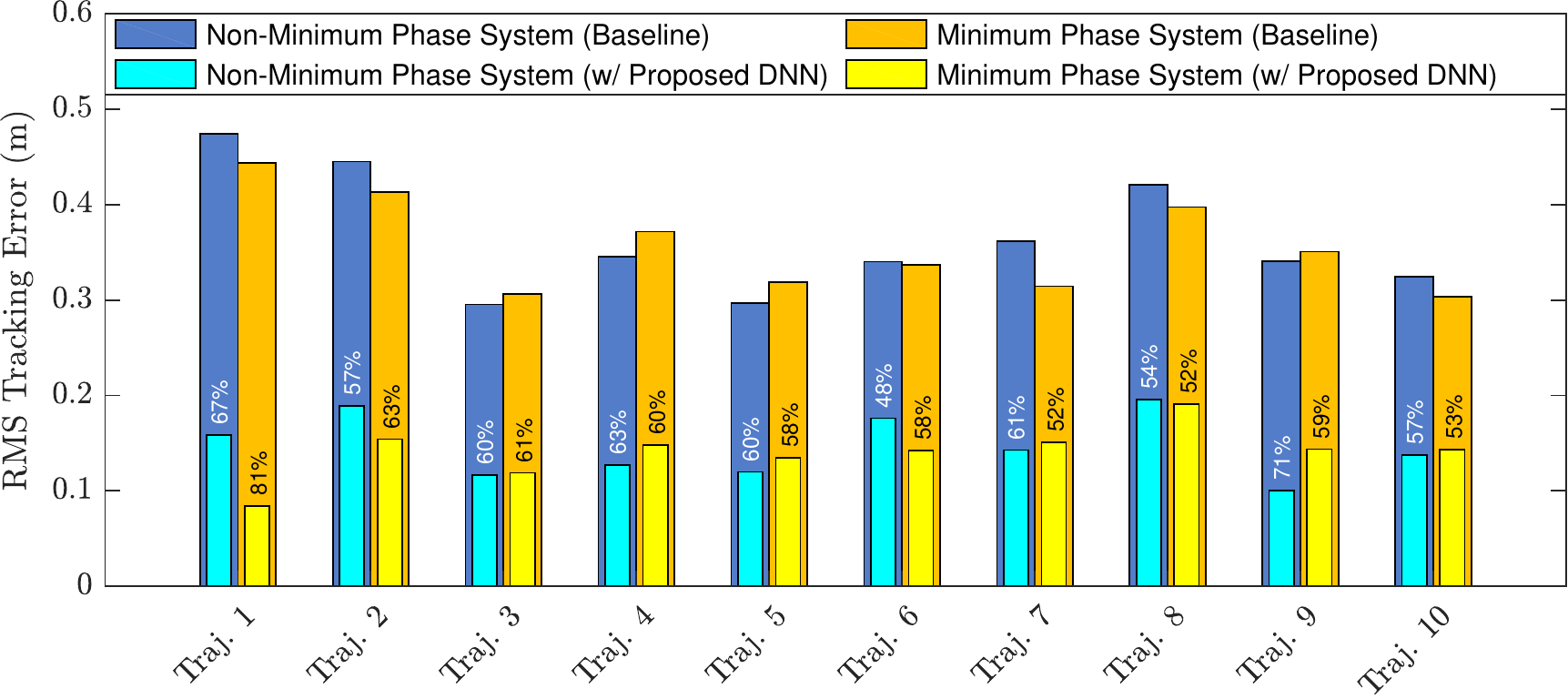}
\vspace{-0.8em}
\caption{\majorChange{RMS tracking errors on 10 hand-drawn test trajectories (shown in Fig.~\ref{fig:testTrajectories}) for the modified non-minimum phase quadrotor system and the original minimum phase quadrotor system. The percentage above each bar indicates the error reduction achieved by the addition of the proposed DNN module. On average, the DNN modules lead to approximately 60\% error reductions for both the non-minimum phase and minimum phase systems.}}
\label{fig:rms_reduction_quadrotor}
\vspace{-1em}
\end{figure}

Fig.~\ref{fig:dsl_tracking} shows the tracking performance of the proposed approach (M3) on the hand-drawn test trajectory corresponding to that shown in Fig.~\ref{fig:z_trajectory}.
On this hand-drawn test trajectory, the proposed approach reduces the 3-dimensional RMS tracking error by 67\%. The generalizability of the proposed approach is tested on 10 hand-drawn trajectories (Fig.~\ref{fig:testTrajectories}), which are not seen during the training phase. 
Fig.~\ref{fig:rms_reduction_quadrotor} shows a summary of the 3-dimensional RMS errors of the non-minimum phase baseline quadrotor tracking system (dark blue bars) and the system enhanced by the proposed DNN approximate inverse learning (light blue bars). On average, 60\% error reduction is achieved by the proposed DNN module. 
In addition, the dark and light yellow bars in Fig.~\ref{fig:rms_reduction_quadrotor} show that the proposed DNN also effectively enhances the performance of the original minimum phase quadrotor system studied in \cite{DNNimpromptuTrack,zhou-cdc17}. 

\majorChangeFinal{Note that, with the proposed approach, it is expected that the performance enhancement of the DNN module is better for input trajectory frequencies closer to those seen in the training phase; in practice, the DNN inverse module should be trained on a dataset that sufficiently covers the operational space. 
}
%
%



\color{black}
\section{Conclusions and Future Work}
\label{sec:conclusions}
Many robotic systems can exhibit non-minimum phase behaviours; in this paper, we present a learning-based approach to enhance the impromptu tracking performance of non-minimum phase systems.
In our approach, a learning module approximates the inverse of a stabilized baseline system, and the stability of the learning module is ensured through appropriate input selection. As demonstrated with experiments on a pendulum-cart and quadrotor system, 
the proposed approach, requiring only input-output data of the baseline system, leads to significantly better performance as compared to the ZOS approximate inverse, one of the typical model-based approaches in the literature.  
\majorChangeFinal{
A promising direction for future research is to incorporate probabilistic modeling approaches to provide uncertainty and performance enhancement estimates for the learned inverse module. 
}

%




\bibliographystyle{IEEEtran}
\bibliography{IEEEabrv,reference}

\end{document}